\pdfoutput=1

\documentclass[11pt]{article}

\usepackage[final]{acl}

\usepackage{times}
\usepackage{latexsym}

\usepackage[T1]{fontenc}

\usepackage[utf8]{inputenc}

\usepackage{microtype}

\usepackage{inconsolata}

\usepackage{graphicx}
\usepackage{booktabs}
\usepackage{multirow}
\usepackage{enumitem}
\setdescription{leftmargin=0pt}
\usepackage{amsmath}
\usepackage{stfloats}
\usepackage{subcaption}
\usepackage{cleveref}

\newcommand{\methodname}{\textsc{BabyHGRN}}
\definecolor{goodOrange}{RGB}{242,96,53}
\newcommand{\changeurlcolor}[1]{\hypersetup{urlcolor=#1}}  

\title{\methodname{}: Exploring RNNs for Sample-Efficient Pretraining}
\title{\methodname{}: Exploring RNNs for Sample-Efficient Training \\ of Language Models}

\author{Patrick Haller \And Jonas Golde \\
\\
Humboldt-Universität zu Berlin
\\
\texttt{\{patrick.haller.1{\normalfont,} jonas.max.golde{\normalfont,} alan.akbik\}@hu-berlin.de} \And
Alan Akbik \\
}

\begin{document}
\maketitle
\begin{abstract}

This paper explores the potential of recurrent neural networks (RNNs) and other subquadratic architectures as competitive alternatives to transformer-based models in low-resource language modeling scenarios. 
We utilize HGRN2~\citep{hgrn2}, a recently proposed RNN-based architecture, and comparatively evaluate its effectiveness against transformer-based baselines and other subquadratic architectures (LSTM, xLSTM, Mamba). Our experimental results show that \methodname{}, our HGRN2 language model, outperforms transformer-based models in both the 10M and 100M word tracks of the challenge, as measured by their performance on the BLiMP, EWoK, GLUE and BEAR benchmarks. Further, we show the positive impact of knowledge distillation.  
Our findings challenge the prevailing focus on transformer architectures and indicate the viability of RNN-based models, particularly in resource-constrained environments.

\end{abstract}

\section{Introduction}
\label{sec:introduction}

In recent years, natural language processing (NLP) has been revolutionized by transformer-based language models (LMs), like BERT~\citep{devlin2019bertpretrainingdeepbidirectional} or GPT~\citep{brown2020languagemodelsfewshotlearners} and their derivatives, achieving state-of-the-art results~\citep{touvron2023llamaopenefficientfoundation, abdin2024phi3technicalreporthighly} across a wide range of tasks such as machine translation, question answering, and text generation. However, despite their dominance, transformers come with notable limitations: they require extensive training data~\citep{hoffmann2022trainingcomputeoptimallargelanguage} and enormous computational resources, which pose challenges for their use in resource-constrained environments.

These limitations led to an increasing interest in more sample-efficient alternatives and approaches with lower computational requirements~\citep{wang2020linformerselfattentionlinearcomplexity}. The shared task of the BabyLM Challenge~\cite{conll-2023-babylm} systematically explores this trend by training LMs on datasets of limited size (10M words in the "\texttt{strict-small}" and 100M words in the "\texttt{strict}" setup). The resulting models are then evaluated on linguistic and general language understanding tasks.

While most participants in BabyLM Challenge focus on adapting transformers to low-resource settings, we propose revisiting recurrent neural networks (RNNs). Once foundational to sequence modeling tasks~\citep{lample-etal-2016-neural,howard-ruder-2018-universal}, RNNs have been largely overshadowed by transformers due to their sequential nature which does not easily allow for parallelization. 

\noindent 
\textbf{Potential of RNN-architectures.}
In this paper, we investigate whether the inductive biases of RNN architectures, such as their sequential processing and memory states, provide advantages in data-constrained settings. This question is especially relevant given that state-of-the-art transformer models depend on quadratic self-attention, which requires calculating the inner product between all tokens. In particular, we investigate the potential of the HGRN2~\citep{hgrn2}, a novel subquadratic RNN-based architecture based on hierarchical gating. We train our model using knowledge distillation~\citep{kd} and evaluate our approach, \methodname{}, against state-of-the-art transformer models and other efficient RNN architectures (e.g. xLSTM~\citep{xlstm} or Mamba~\citep{mamba}). Our experiments demonstrate that our resulting model yields better performance compared to both transformer-based and other RNN-based architectures.


\noindent We summarize our contributions as follows:
\begin{enumerate}
    \item We conduct an exploratory evaluation of transformer-based and other RNN-based architectures (HGRN2, LSTM, xLSTM, Mamba), contributing to the ongoing research on sample-efficient language modeling.
    \item We present a comprehensive evaluation of our proposed HGRN2 language model \methodname{}. We show the impact of knowledge distillation and the choice of dataset.
    \item We release all code, datasets, and experimental setups to the research community to facilitate reproducibility and further research\footnote{\changeurlcolor{goodOrange}\url{https://github.com/HallerPatrick/BabyLM-2024}}.
\end{enumerate}

Our results show that \methodname{} outperforms transformer-based baselines on both tracks of the BabyLM challenge.

\begin{table*}[t]
\begin{minipage}{.5\linewidth}
\centering
\begin{tabular}{lrr}
\toprule
Dataset & Count & Ratio (\%) \\
\midrule
Pile-CC & 4,900,155 & 49.00 \\
OpenWebText2 & 3,078,791 & 30.79 \\
FreeLaw & 946,382 & 9.46 \\
USPTO Backgrounds & 261,159 & 2.61 \\
Wikipedia (en) & 187,094 & 1.87 \\
PubMed Central & 142,698 & 1.43 \\
PubMed Abstracts & 118,427 & 1.18 \\
Others & 365,188 & 3.65 \\
\midrule
Total & 9,999,894 & \\
\bottomrule
\end{tabular}
    \end{minipage}%
    \begin{minipage}{0.52\linewidth}
      \centering
\begin{tabular}{lrr}
\toprule
Dataset & Count & Ratio (\%) \\
\midrule
Pile-CC & 49,214,555 & 49.21 \\
OpenWebText2 & 30,344,790 & 30.34 \\
FreeLaw & 9,471,436 & 9.47 \\
USPTO Backgrounds & 2,519,390 & 2.52 \\
Wikipedia (en) & 1,855,709 & 1.86 \\
PubMed Central & 1,449,273 & 1.45 \\
PubMed Abstracts & 1,175,838 & 1.18 \\
Others & 3,968,870 & 3.97 \\
\midrule
Total & 99,999,861 & \\
\bottomrule
\end{tabular}

    \end{minipage} 

\caption{Composition of the \textbf{10M} (left table) and \textbf{100M} (right table) word datasets (word counts and ratio per domain) we created from the \textsc{Pile} to train \methodname{}.}
\label{tab:pile_tokens_count}
\end{table*}

\section{\methodname{}}
We utilize HGRN2 as our backbone architecture with a hidden size of 2048 and 18 layers, resulting in a total parameter count of ~330M. We train our model either with (\textit{1}) the default dataset of the BabyLM Challenge or (\textit{2}) a sub-sampled split of ThePile~\citep{pile}. Further, we employ knowledge distillation training using a teacher-student setup. In the following, we will discuss the details of our design choices.

\subsection{Training Dataset}
\label{sec:pile}

We curate our own training datasets for the \texttt{strict} and \texttt{strict-small} tracks by sub-sampling the Pile dataset (see Table~\ref{tab:pile_tokens_count}). The Pile consists of 22 smaller datasets that cover a variety of domains, including books, web pages, scientific literature, and programming code. 
The main motivation behind choosing the Pile dataset is its diverse composition, which may offer several advantages for language model training. Approximately 14\% of the original BabyLM dataset consists of child-related text (e.g., the Children's Book Test~\citep{hill2016goldilocksprinciplereadingchildrens}, Children's Stories Text Corpus\footnote{\changeurlcolor{goodOrange}\url{https://www.kaggle.com/datasets/edenbd/children-stories-text-corpus}}, and CHILDES project~\citep{childes}), which may limit its generalizability across diverse domains. In contrast, the broader scope of the Pile dataset could improve resilience in zero-shot tasks and potentially enhance adaptability for fine-tuning on specific areas of interest.

We create the splits by randomly sampling from each chosen subset until we reached the pre-defined thresholds. We depict details on our selected subsets and corresponding word counts in~\Cref{tab:pile_tokens_count}.

To minimize computational overhead, we concatenate all samples and segment them into uniform chunks of 512 tokens. Subsequently, each input sample is tokenized using Byte-Pair Encoding (BPE), employing a vocabulary size of 16,000 tokens. We chose the \textit{BabyLlama} tokenizer provided with the baseline models by the organizers\footnote{\changeurlcolor{goodOrange}\url{https://huggingface.co/babylm/babyllama-100m-2024}}.

\subsection{Training Objectives}
\label{sec:objectives}
We use standard next-token prediction as the language modeling task and employ token-level cross-entropy loss for training our models. For a sequence of tokens $x = (x_1, ..., x_N)$, the loss is calculated as:
$$\mathcal{L}(\theta) = - \frac{1}{N} \sum_{i=1}^N \log P(x_i|x_1, ..., x_{i-1}; \theta)$$
where $\theta$ represents the model parameters and $P(x_i \mid x_1, ..., x_{i-1}; \theta)$ is the probability the model assigns to the i-th token given all previous tokens.

We further improve our model through knowledge distillation~\citep{kd_model_compression,kd}, where we train a second HGRN2 model (student) using predictions from our initially trained model (teacher). While knowledge distillation traditionally transfers knowledge from larger to smaller models, using same-sized teacher and student models has proven effective in recent work - notably in the previous BabyLM Challenge where an ensemble of teachers was used for knowledge transfer~\citep{babyllama}.

The training process for the student model incorporates an additional loss term based on soft labels produced by the teacher model. The total loss function for the student model can be expressed as:
$$\mathcal{L}_\text{total} = (1 - \alpha) \mathcal{L}_\text{CE} + \alpha \mathcal{L}_\text{KD}$$
where $\mathcal{L}_{\text{CE}}$ is the standard cross-entropy loss for the student model, $\mathcal{L}_\text{KD}$ is the knowledge distillation loss, and $\alpha$ is a hyperparameter that balances the two loss terms.

In our implementation, the knowledge distillation loss $\mathcal{L}_\text{KD}$ is calculated using the Kullback-Leibler divergence between the probability distributions of the teacher and student models:
$$\mathcal{L}_\text{KD} = \text{KL}(\sigma(z_t) || \sigma(z_s))$$
where $z_t$ and $z_s$ are the output logits of the teacher and student model respectively. And $\sigma(z)$ is the softmax function applied to the logits $z$.



\subsection{Training Details}



\noindent
For fine-tuning on the (Super)Glue tasks, we follow the provided hyperparameters by the shared task organizer (see Appendix~\ref{sec:finetune_hp}). Except for the \textit{WSC} tasks, which had unusually low scores. We used a maximum of 20 epochs, a patience of 6 epochs and a learning rate of $1\times 10^{-5}$ for our final submission models.

\noindent
\textbf{Software.}
For training our model we use the \texttt{Pytorch}~\citep{pytorch2} library. Relevant metrics are logged with \texttt{Weights and Biases}~\citep{wandb}. We use HuggingFace \texttt{datasets}~\citep{hf_datasets} library for dataset loading and subsampling. All relevant models were either directly 
imported with the \texttt{transformers}~\citep{hf_transformers} library or implemented as a custom model. For the HGRN2 model we used the FLA~\citep{fla} library.

\noindent
\textbf{Hardware.}
All models were trained with the \texttt{torch.distributed} package in data-parallel mode. Models were trained on 4 RTX A6000 49GB graphics cards on one node.

\section{Empirical Evaluation}

In~\Cref{sec:babylm_eval_datasets}, we shortly present the evaluation benchmarks of the BabyLM Challenge and the BEAR knowledge probe. In~\Cref{sec:preliminary_experiments,sec:learning_dynamics,sec:custom_dataset}, we evaluate \methodname{} compared with other efficient RNN architectures, its training dynamics, and the influence of different datasets. Finally, in~\Cref{sec:knowledge_distillation}, we evaluate \methodname{} using knowledge distillation.

\subsection{Evaluation Datasets} \label{sec:babylm_eval_datasets}
The BabyLM challenge covers three benchmarks: BLiMP~\cite{blimp}, EWoK~\citep{ewok}, and parts of GLUE~\citep{glue} and SuperGLUE~\citep{superglue}, respectively. These benchmarks are designed to assess language model performance such as grammatical knowledge or complex reasoning tasks. Additionally, we include the BEAR probe~\citep{wiland-etal-2024-bear} to evaluate factual knowledge capabilities.

\noindent\textbf{BLiMP}~(Benchmark of Linguistic Minimal Pairs) is an English zero-shot benchmark evaluating the grammatical knowledge of language models. It has 67 sub-tasks, each focusing on a specific syntactic or semantic phenomenon. Specifically, the dataset contains pairs of sentences and the model is tasked to differentiate which of the sentences is grammatically correct. Further, we consider the hidden task \textit{"BLiMP Supplement"} of the 2023 BabyLM Challenge \citep{conll-2023-babylm}.

\noindent\textbf{EWoK}~(Elements of World Knowledge) evaluates basic world knowledge in language models. This cognition-inspired approach tests whether language models can identify plausible contexts given different fillers. EWoK was introduced as the hidden task for the 2024 BabyLM Challenge.

\noindent\textbf{GLUE}~(General Language Understanding Evaluation) is a multi-task benchmark evaluating natural language understanding systems. It contains nine tasks such as sentiment analysis, question answering, or textual entailment. As models began to surpass human performance on several GLUE tasks, SuperGLUE was introduced as an extension, including more challenging tasks.

\noindent\textbf{BEAR}~\citep{wiland-etal-2024-bear} tests relational knowledge in language models using 7,731 instances over 60 relations. BEAR compares the models' log-likelihood for different factual statements of which only one is true. We leverage the implementation by~\citet{ploner2024lmpubquiz} to conduct the BEAR probing experiments.

\begin{table*}[ht]
\centering
\begin{tabular}{lccccccr}
\toprule
Model & \#Params & Epoch & BLiMP & BLiMP-Supp. & EWoK & Macro-Avg. \\
\midrule
Transformer   & 360M & 4 & 62.64 & 54.86 & 50.48 &  55.99 \\
\midrule
LSTM        & 300M  & 5 & 62.27 & 51.63 & 50.48 & 54.79  \\
Mamba       & 350M  & 2 & 64.44 & 55.39 & 50.39 &  56.74 \\
xLSTM       & 340M  & 3 & 64.66 & 56.72 & 49.48 & 56.95 \\
HGRN2       & 360M  & 4 & 67.05 & 55.69 & 49.88 & \textbf{57.54} \\
\bottomrule
\end{tabular}
\caption{Results from training on the 10M word corpus, comparing various RNN architectures to a Transformer-based model (LLaMA architecture). Each model was trained for 5 epochs, with evaluations after each epoch, and the best-performing model was selected.}
\label{tab:comparison}
\end{table*}

\subsection{Experiment 1: RNN Architecture Selection} \label{sec:preliminary_experiments}

\begin{table}[t]
\centering
\begin{tabular}{lr}
\toprule
Hyperparameter & Value \\
\midrule
Epochs          & 3 \\
Batch Size      & 64 \\
Learning Rates  & \{1e-3, 1e-4, 1e-5, 1e-6\} \\
Optimizer       & Adam \\
Sequence Length & 512 \\
Max Grad Norm   & 1.0 \\
LR Scheduler    & Linear \\
\bottomrule
\end{tabular}
\caption{Pretraining hyperparameters used for all models and experiments.}
\label{tab:hyperparameters}
\end{table}

Our first experiment compares the HGRN architecture with other RNN-based and transformer architectures. Specifically, we compare HGRN2, the vanilla LSTM, xLSTM, Mamba, and a Transformer baseline.

\noindent 
\textbf{Experimental setup.}
We select configurations such that all architectures have a similar parameter count of 300 to 360 million. We use the configurations as as originally proposed for xLSTM, Mamba, and HGRN2. For the decoder-only transformer, we use the LLaMA~\citep{touvron2023llamaopenefficientfoundation} model and follow the Pythia~\citep{biderman2023pythiasuiteanalyzinglarge} 410M model configuration with 22 hidden layers. For the vanilla LSTM, we set the hidden size to 4096 with two layers to match the parameter count of the other architectures. We refer to~\Cref{sec:model_configurations} for a detailed overview of all configurations.

For each architecture, we perform learning rate selection for all considered architectures by executing a grid search over commonly used learning rates (\{1e-3, 1e-4, 1e-5, 1e-6\}). We train each model for 5 epochs on the \texttt{strict-small} dataset of the BabyLM challenge. Further, we do not employ any knowledge distillation and train all LMs using the next-token prediction objective. We report results on the zero-shot benchmarks of BabyLM, namely BLiMP and EWoK, together with their best hyperparameter configuration.



\noindent 
\textbf{Results.}
Table~\ref{tab:comparison} shows the number of parameters of each considered architecture and the results achieved during the exploration phase on the zero-shot benchmarks\footnote{We report the complete results of the parameter sweep in~\Cref{sec:lr_sweep}.}. We find that the HGRN2 exhibits the best performance, closely followed by xLSTM and Mamba. Both outperform the transformer model, suggesting that these architectures offer advantages in low-resource scenarios. 
The standard LSTM, serving as a baseline for classical RNN architectures and performs worse than the transformer model. 
Further, we observe that all architectures perform best using a learning rate of $1e^{-3}$.


\subsection{Experiment 2: Learning Dynamics of HGRN2} \label{sec:learning_dynamics}

To better understand the learning dynamics of the selected HGRN2 architecture, we investigated how its zero-shot performance on the BabyLM benchmark changes over the epochs during training.

\noindent 
\textbf{Experimental setup.} We re-use the best performing hyperparameters from~\Cref{sec:preliminary_experiments}. After each epoch, we evaluate on BLiMP, BLiMP Supp. and EWoK.

\begin{figure}[t]
    \centering
    \includegraphics[width=1.1\linewidth]{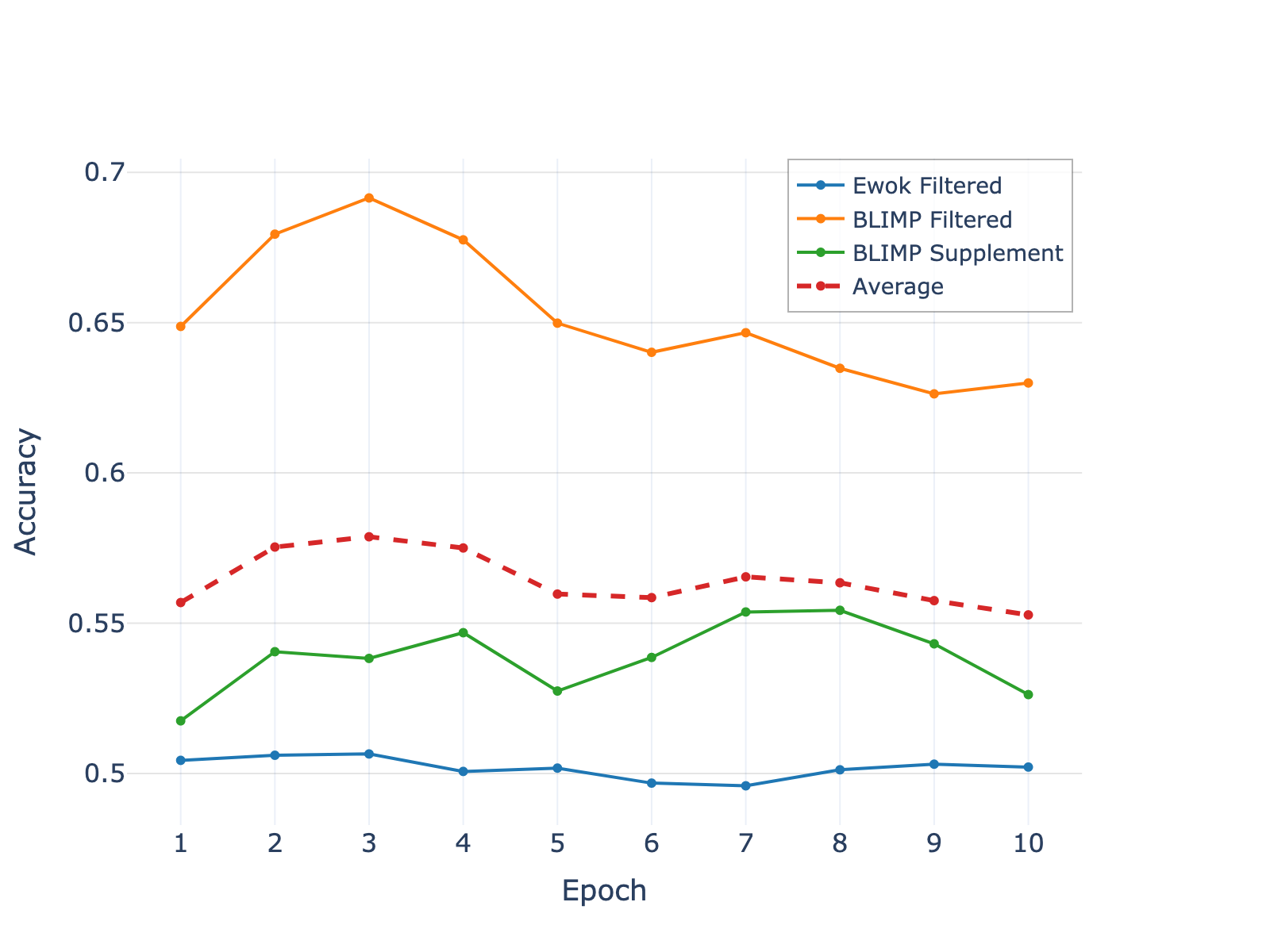}
    \caption{Performance evaluation of epochs of pretraining, with the macro average at epoch 3 being the highest.}
    \label{fig:eval_epochs}
\end{figure}

\noindent 
\textbf{Results.} The results of this experiment are illustrated in~\Cref{fig:eval_epochs}. Our analysis reveals early peaks in performance on BLiMP and EWoK and a later peak on BLiMP Supplement. This finding indicates that HRGN2 initially captures certain linguistic patterns from the limited training data, although the gains over random baseline are modest. Further iterations yield only incremental improvements, which may point to constraints in the model’s ability to leverage the available data fully. 



\subsection{Experiment 3: Impact of Training Dataset} \label{sec:custom_dataset}
In this experiment, we evaluate the impact of the choice of training data. We compare models trained over the default BabyLM dataset to models trained using our custom dataset derived from the Pile (see \cref{sec:pile}). 

\noindent
\textbf{Experimental Setup.}
We re-use our chosen hyperparameter configuration for the HGRN2 architecture from~\Cref{sec:learning_dynamics} and train two models on (\textit{1}) our derived Pile subset and (\textit{2}) on the default BabyLM dataset. We train models for 5 epochs, evaluate after each epoch, and report results of the best performing model. In this experiment we include both the 10M and 100M word datasets for a full comparison.

\noindent
\textbf{Results.}~\Cref{tab:dataset_comparison} summarizes the performance across all benchmarks. For the 10M word track, the HGRN2 model trained on our derived dataset shows modest gains on BLiMP, EWoK, and BEAR, but underperforms on the BLiMP-Supplemental subset ($\downarrow$3.47 pp). This suggests that at smaller data scales, our dataset may lack certain syntactic structures present in the original BabyLM dataset. Furthermore, given the limited dataset size in the 10M word track, these numbers may lack statistical significance.

In contrast, the 100M word track demonstrates consistently stronger performance across all metrics, with particularly notable improvements on BLiMP ($\uparrow$3.45 pp) and BEAR ($\uparrow$1.21 pp). Indicating that our dataset selection strategy enhances the model's ability to acquire both syntactic and factual knowledge when given sufficient training data.

\begin{table}[t]
    \centering
    \resizebox{\columnwidth}{!}{%
    \begin{tabular}{lccc|c}
    \toprule
    Dataset & BLiMP & BLiMP-Supp. & EWoK & BEAR\\
    \midrule
     BabyLM - 10M & 67.05 & 55.69 & 49.88 & 5.29 \\
     Ours - 10M & 67.49 & 52.22 & 50.62 & 5.36 \\
     \midrule
     BabyLM - 100M & 69.44 & 55.56 & 50.31 & 6.17 \\
     BabyHGRN - 100M - Epoch 1 & \textbf{72.89} & \textbf{57.43} & 50.61 & \textbf{7.38} \\
    \bottomrule
    \end{tabular}}
    \caption{Zero-shot evaluation results comparing HGRN2 models trained on the BabyLM dataset versus our proposed Pile subset. Both models were trained with a learning rate of $1\times10^{-3}$. All metrics are reported as percentages.}
    \label{tab:dataset_comparison}
\end{table}

\subsection{Experiment 4: \methodname{} With Knowledge Distillation} \label{sec:knowledge_distillation}




\begin{table*}[ht]
\setlength{\tabcolsep}{5pt} 
\centering
\begin{tabular}{lccccr|r}
\toprule
   & BLiMP & BLiMP-Supp. & EWoK & SuperGLUE & Average & BEAR \\
 \midrule
 BabyLlama & 69.8 & 59.5 & 50.7 & 63.3 &  60.8 & 5.4 \\
 LTG-BERT  & 60.6 & 60.8 & 48.9 & 60.3 &  57.7 & 5.7 \\
 \midrule
 BabyHGRN$_\textit{ce}$   \textit{(ours)}                & 69.4 & 55.6 & 50.7 &  63.0 & 59.7 & 5.6 \\
 BabyHGRN \textit{(ours)}  & 72.1 & 58.6 & 51.3 & 65.8 & \textbf{63.3} & 7.5 \\
 \bottomrule
\end{tabular}
\caption{Evaluation results for the \textbf{10M words} track ("strict-small"). The BabyLM score is computed as a macro average over four datasets (BLiMP, BLiMP Supp., EWoK and SuperGLUE) but note that the macro average may not be a representative overall score for each model, since the datasets are of widely varying size (e.g. the BLiMP supplements is only 7\% in size compared to the BLiMP). We additionally include the BEAR score for comparison and evaluation of factual knowledge.}
\label{tab:overall_results_10m}
\end{table*}

\begin{table*}[ht]
\setlength{\tabcolsep}{5pt} 
\centering
\begin{tabular}{lccccr|r}
\toprule
   & BLiMP & BLiMP-Supp. & EWoK & SuperGLUE & Average & BEAR \\
 \midrule
 BabyLlama  & 73.1 & 60.6 & 52.1 & 69.0 & 63.7 & 8.5 \\
 LTG-BERT & 69.2 & 66.5 & 51.9 & 68.4 & 64.0 & 8.2 \\
 \midrule
 BabyHGRN$_\textit{ce}$ \textit{(ours)} & 74.5 & 59.1 & 52.88 & 69.1 & 63.9 & 13.5 \\
 BabyHGRN  \textit{(ours)}  & \textbf{77.5} & 58.5 & 51.6 & \textbf{70.7} &  \textbf{64.9} & \textbf{13.6} \\
 \bottomrule
\end{tabular}
\caption{Evaluation results for the \textbf{100M words} track ("strict"). The BabyLM score is computed as a macro average over four datasets (BLiMP, BLiMP Supp., EWoK and SuperGLUE). We additionally include the BEAR score for comparison and evaluation of factual knowledge.}
\label{tab:overall_results_100m}
\end{table*}


Based on the exploratory experiments of the previous subsections, we selected the HGRN2 model trained on our proposed dataset for the BabyLM challenge. We furthermore apply knowledge distillation as outlined in Section~\ref{sec:objectives} to our final model. We refer to this model as \methodname{}.


In this section, we evaluate \methodname{} using knowledge distillation learning and compare it with two baselines (BabyLlama and LTG-BERT) and a \methodname{} version using only the cross entropy objective. We denote the ablationm odel as BabyHGRN$_\textit{ce}$.

\noindent
\textbf{Hyperparameters.}
We increase the model size in accordance with scaling laws for language models~\citep{kaplan2020scalinglawsneurallanguage} from 360M to 1.0B. We reduce the learning rate from $1\times10^{-3}$ to $4\times10^{-4}$ accordingly, following the configuration found in~\Cref{sec:preliminary_experiments,sec:learning_dynamics}. Empirical work~\citep{kaplan2020scalinglawsneurallanguage,hoffmann2022trainingcomputeoptimallargelanguage} suggests that lower learning rates in larger models help mitigate instabilities during training, promoting smoother convergence and more efficient use of computational resources.


\subsubsection{Results}

\Cref{tab:overall_results_10m} and~\Cref{tab:overall_results_100m} summarize our experimental results for the 10M and 100M word tracks, respectively.

\noindent 
\textbf{HGRN2 outperforms baselines.} Most importantly, we find that our 
HGRN2 models show competitive performance across both the 10M and 100M word tracks of the BabyLM challenge. On the 10M words track, BabyHGRN achieves an overall macro average of 63.3\% ($\uparrow$2.5 pp vs. BabyLlama).  
As~\Cref{tab:overall_results_10m} shows, BabyHGRN particularly outperforms the baselines on the BLiMP ($\uparrow$2.4 pp vs. BabyLlama) and SuperGLUE ($\uparrow$2.5 pp vs. BabyLlama) tasks, and significantly improves the BEAR score ($\uparrow$1.8 pp vs. LTG-BERT).

On the 100M words track (refer to~\Cref{tab:overall_results_100m}), BabyHGRN outperforms the baselines with a marco average of 64.9\% ($\uparrow$0.9 pp vs. LTG-BERT), though the improvement is not as pronounced as in the more data-constrained 10M scenario. Here, BabyHGRN improves in particular the BLiMP ($\uparrow$4.4 pp vs. LTG-BERT) and SuperGLUE ($\uparrow$1.7 pp vs. BabyLlama) tasks, but falls short on BLiMP-Supplement ($\downarrow$7.4 pp vs. LTG-BERT)\footnote{Detailed results for BLiMP, BLiMP-Supplement, EWoK and (Super)Glue are provided in~\Cref{sec:final_scores}.}.

\noindent 
\textbf{Knowledge distillation is helpful.} We also note that our knowledge distillation approach significantly improves performance of \methodname{}, compared to the distillation-free approach BabyHGRN$_{ce}$. As~\Cref{tab:overall_results_10m,tab:overall_results_100m} show, BabyHGRN outperforms both, BabyLlama and LGT-BERT, baselines. Further, we observe \methodname{} outperforms BabyHGRN$_{ce}$ by 5.3 pp on average in the data-constrained 10M setting, confirming the usefulness of distillation losses in such settings.

\noindent
\textbf{\methodname{} is better at learning factual knowledge.} 
While the accuracy on BEAR is relatively low across all settings (compared to state-of-the-art models such as LLaMA-3 with 68.6), we observe that \methodname{} strongly outperforms transformer-based baselines in data-restricted settings. For instance, BEAR shows a pronounced difference between BabyHGRN and BabyHGRN$_{ce}$ on the 10M track, and a large difference between the HGRN models and the baselines on the 100M track. We primarily attribute this improvement to the use of our custom dataset.

\section{Related Work}

In recent years, there has been a resurgence of interest in recurrent neural network (RNN) architectures for sequence modeling, particularly in the context of large language models (LLMs). This renewed focus has led to the development of several RNN-based architectures that aim to combine the efficiency of recurrent models with the expressiveness of more complex architectures like transformers.

\begin{description}
\item[HGRN and HGRN2]The Hierarchically Gated Recurrent Neural Network (HGRN) \citep{hgrn} introduces a novel gating mechanism that allows for more effective modeling of long-term dependencies. The key innovation of HGRN is its hierarchical structure, in which forget gates have monotonically increasing lower bound values from bottom layers to upper layers. This design enables lower layers to model short-term dependencies while upper layers capture long-term relationships in the data.
HGRN achieves efficient training by reformulating its recurrent computation as a parallel scan operation to enable parallelization across sequence length while maintaining linear time complexity.

Building upon HGRN, \citet{hgrn2} introduced HGRN2 which further enhances the capabilities of gated linear RNNs. HGRN2 addresses some limitations of its predecessor by incorporating a state expansion mechanism. This innovation significantly increases the recurrent state size without introducing additional parameters, leading to improved expressiveness.

\item[xLSTM] Another recently proposed RNN-based architecture is the Extended Long Short-Term Memory (xLSTM) \citep{xlstm}. xLSTM builds upon the classical LSTM \citep{lstm} by introducing two key modifications: exponential gating and modified memory structures. The exponential gating mechanism allows the model to revise storage decisions more effectively, addressing a key limitation of traditional LSTMs. xLSTM introduces two variants: sLSTM with a scalar memory and new memory mixing technique, and mLSTM with a matrix memory and covariance update rule, which is fully parallelizable.
The xLSTM approach demonstrates strong performance across various modalities, including language, vision~\citep{vision_xlstm, vision2_xlstm}, and audio~\citep{audio_xlstm}, while maintaining linear scaling in sequence length and efficient inference.

The \textbf{Mamba} architecture~\citep{mamba} improves on state space models (SSMs) by introducing selective state spaces. Building on structured SSMs~\citep{structured_ssm}, Mamba achieves linear-time sequence processing through input-dependent SSM parameters, enabling selective information propagation across sequences. This mechanism is conceptually similar to gating in classical RNNs~\citep{lstm} while maintaining modern computational benefits. The architecture consists of repeated blocks that combine selective SSMs with feed-forward components, in contrast to more complex predecessors like H3~\citep{h3} and Hyena~\citep{hyena}. Though attention-free, Mamba matches or exceeds Transformer performance~\citep{vaswani2023attentionneed} across various domains. Its recurrent computation pattern eliminates the need for attention caches during inference, leading to 5× faster inference compared to similar-sized Transformers. This combination of linear scaling and efficiency, without sacrificing model quality, makes Mamba a significant development in sequence modeling.

\end{description}

The development of HGRN2, xLSTM, and Mamba is part of a broader trend in revisiting and improving RNN architectures \citep{rwkv, retnet}.

\section{Conclusion}
We presented BabyHGRN, an RNN-based language model that utilizes the HGRN2 architecture. Our experimental results on the evaluation datasets of the BabyLM Challenge and the BEAR probe indicate that BabyHRGN is competitive. Indeed, despite relatively little hyperparameter optimization, our approach significantly outperforms strong transformer-based baselines on the evaluation datasets. 

Revisiting our research question posed in Section~\ref{sec:introduction}, we conclude that RNN-based language models are indeed competitive in low-resource language modeling scenarios. Based on these results, we believe that advanced RNN-based architectures such as HGRN and Mamba may hold promise for research in sample-efficient language modeling. Accordingly, future work could explore further optimizations of the underlying RNN architectures, investigate their performance on a broader range of tasks, and examine their scalability to larger datasets and model sizes.

\clearpage
\section*{Limitations}

Our experiments with HGRN2 in the BabyLM Challenge demonstrate the competitiveness of RNN-based models with transformers in low-resource scenarios.
However, while we find our results to be promising, it's important to acknowledge that there are several avenues for optimization that we have yet to explore:

\begin{description}
    \item[Dataset sampling] The dataset we used to train BabyHGRN was produced using a naive random sampling of the \textsc{Pile} dataset. More sophisticated approaches, such as importance sampling specialized for downstream tasks, would likely yield better results, especially if optimized for the tasks BabyLM evaluates on. In our work, we refrained from such "dataset engineering" and focused solely on a comparison of different RNN architectures.
    \item[Model configurations] We utilized the configurations provided by the authors of HGRN2 and xLSTM. Further experimentation with different architectures and hyperparameters for the low-resource scenario could well lead to improved performance of these models.
    \item[Context length] Optimizing the context length for our specific tasks and data could potentially enhance the model's capabilities. Work from previous years challenge~\citep{context_size_1, context_size_2} suggests that a smaller context size improves performance on all benchmarks.
    \item[Knowledge distillation] As previously discussed, we only implemented a basic knowledge distillation approach to train BabyHGRN. More sophisticated techniques, such as those employed by \citet{babyllama} could further boost performance.

\end{description}

Our work thus serves as a proof of concept, demonstrating that RNNs can be competitive with transformers in this domain, while leaving room for further advancements.

\section*{Acknowledgements}
We thank all reviewers for their valuable comments. Alan Akbik and Patrick Haller are supported by the Deutsche Forschungsgemeinschaft (DFG, German Research Foundation) under Emmy Noether grant “Eidetic Representations of Natural Language” (project number 448414230).
Alan Akbik is furthermore supported under Germany’s Excellence Strategy "Science of Intelligence" (EXC 2002/1, project number 390523135).
Jonas Golde is supported by the Bundesministerium für Bildung und Forschung (BMBF) as part of the project "FewTuRe" (project number 01IS24020).

\bibliography{anthology_1,custom}

\appendix


\section{Finetune Hyperparameters}\label{sec:finetune_hp}

\begin{figure}[h]
    \centering
    \begin{tabular}{lr}
    \toprule
    Hyperparameter & Value \\
    \midrule
    Initial learning rate & 5e-5 \\
    Batch size & 64 \\
    Maximum epochs & 10 \\
    Evaluate every (epochs) & 1 \\
    Patience & 3\\
    \bottomrule
    \end{tabular}
    \caption{Default hyperparameters for fine-tuning on the (Super)Glue tasks.}
    \label{fig:enter-label}
\end{figure}
\newpage

\section{Model Configurations}\label{sec:model_configurations}
\begin{table}[h]
\centering
\small  
\begin{tabular}{lr}
\toprule
\textbf{Transformer} & Value \\
\midrule
Hidden Size       & 1024 \\
Intermediate Size & 4096 \\
Hidden Layers     & 22 \\
Attention Heads   & 32 \\
\midrule
\textbf{LSTM} & Value \\
\midrule
Hidden Size     & 9120 \\
Embedding Size  & 512 \\
LSTM Layers     & 2 \\
Dropout         & 0.1 \\
\midrule
\textbf{Mamba} & Value \\
\midrule
Hidden Size       & 1024 \\
Intermediate Size & 2048 \\
Hidden Layers     & 48 \\
State Size       & 8 \\
\midrule
\textbf{xLSTM} & Value \\
\midrule
Embedding Size  & 1024 \\
Num Blocks      & 48 \\
mLSTM Heads     & 4 \\
Ratio           & [1:0] \\
\midrule
\textbf{HGRN2 - 360M} & Value \\
\midrule
Hidden Size     & 1024 \\
Layers          & 26 \\
Hidden Ratio    & 4 \\
Expand Ratio    & 128 \\
\midrule
\textbf{HGRN2 - 1.2B} & Value \\
\midrule
Hidden Size     & 2048 \\
Layers          & 18 \\
Hidden Ratio    & 4 \\
Expand Ratio    & 128 \\
\bottomrule
\end{tabular}
\caption{Complete list of model configurations.}
\end{table}

\newpage
\section{Learning Rate Parameter Sweep}\label{sec:lr_sweep}
\begin{figure}[h]
    \centering
    \includegraphics[width=0.40\textwidth]{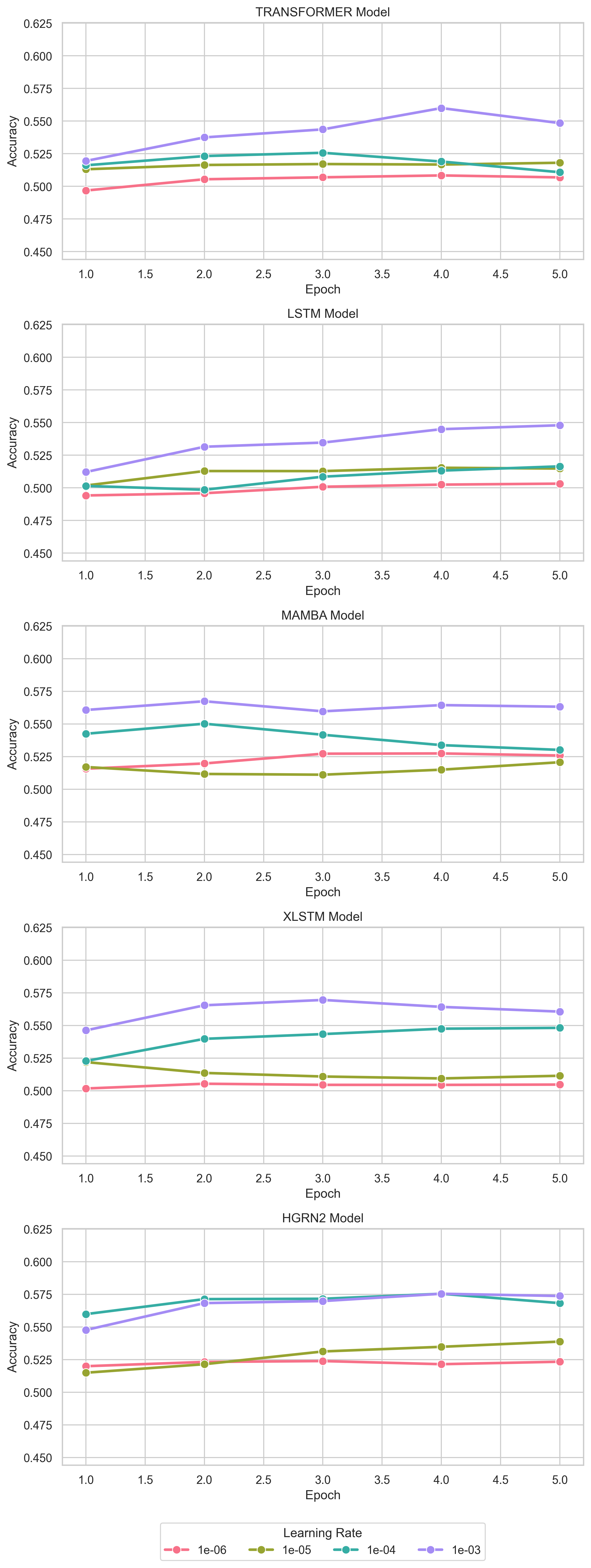}
    \caption{Evaluation results of learning rate sweep over different architectures. Scores are reported as the macro average over the three zero-shot benchmarks BLiMP, BLiMP-Supplement and EWoK.}
    \label{fig:enter-label}
\end{figure}

\newpage
\section{Final BabyLM Evaluation Scores}\label{sec:final_scores}

We provide detailed scores of all SuperGLUE, BLiMP-Supplement and EWoK tasks in Tables~\ref{tab:results_strict_small}, \ref{tab:model_comparison} and \ref{tab:ewok_scores}. Due to the large number of subtasks in BLiMP, we will make the scores accessible though our Github repository: \changeurlcolor{goodOrange}\url{https://github.com/HallerPatrick/BabyLM-2024}.

{\renewcommand{\arraystretch}{1.15}
\begin{table*}[h]
\resizebox{\textwidth}{!}{%
\begin{tabular}{@{}l@{\hspace{1.0em}}ccccccccccccc@{\hspace{1.0em}}}
\toprule
\multicolumn{1}{@{}l}{\multirow{2}{*}{\vspace{-1.5em}\textbf{Model (variant)}}} &
\multicolumn{12}{@{}c@{}}{\hspace{-3em}~~\textbf{SuperGLUE}\hspace{2em}} &
\\

\multicolumn{1}{l}{} &
\footnotesize{\textbf{BoolQ}} &
\footnotesize{\textbf{CoLA (MCC)}} &
\footnotesize{\textbf{MNLI}} &
\footnotesize{\textbf{MNLI-MM}} &
\footnotesize{\textbf{MRPC (F1)}} &
\footnotesize{\textbf{MulitRC}} &
\footnotesize{\textbf{QNLI}} &
\footnotesize{\textbf{QQP (F1)}} &
\footnotesize{\textbf{RTE}} &
\footnotesize{\textbf{SST-2}} &
\footnotesize{\textbf{WSC}} &
\footnotesize{\textbf{Average}} \\

\midrule
\multicolumn{13}{@{}c@{}}{\vspace{0.4em}\raisebox{-0.4em}{\footnotesize{Strict-small Track (10M Words)}}} \\
BabyLlama$_\text{baseline}$                &
65.0 &
2.2 &
72.4 &
74.2 &
82.0 &
60.1 &
82.8 &
83.6 &
49.6 &
86.2 &
38.5 &
63.3 \\
LTG-BERT$_\text{baseline}$                &
68.8 &
0.0 &
68.9 &
68.9 &
82.2 &
58.5 &
76.5 &
34.2 &
58.3 &
85.1 &
61.5 &
60.3 \\
\midrule
BabyHGRN$_\text{ce}$                &
63.8 &
19.1 &
\textbf{68.7} &
\textbf{68.7} &
82.5 &
63.4 &
64.7 &
79.9 &
58.9 &
85.5 &
38.5 &
63.0 \\
BabyHGRN                &
65.4 &
33.1 &
69.3 &
69.5 &
81.0 &
59.7 &
72.3 &
81.9 &
54.0 &
89.4 &
48.1 &
\textbf{65.8} \\
\midrule
\multicolumn{13}{@{}c@{}}{\vspace{0.4em}\raisebox{-0.4em}{\footnotesize{Strict-small Track (100M Words)}}} \\
BabyLlama$_\text{baseline}$                &
66.1 & 
37.3 & 
75.6 & 
76.2 & 
86.8 & 
62.1 & 
83.1 & 
84.5 & 
60.4 & 
88.3 & 
38.5 & 
69.0 \\
LTG-BERT$_\text{baseline}$                &
61.7 & 
34.6 & 
77.7 &
78.1 & 
83.1 & 
52.6 &
78.2 & 
86.7 & 
46.8 & 
91.5 & 
61.5 &
68.4 \\
\midrule
BabyHGRN$_\text{ce}$                &
64.4 &
39.9 &
\textbf{74.3} &
\textbf{74.3} &
82.8 &
61.4 &
79.9 &
83.1 &
58.9 &
89.6 &
51.6 &
69.1 \\
BabyHGRN                &
64.8 &
40.3 &
74.8 &
75.9 &
81.5 &
61.4 &
81.5 &
84.1 &
58.3 &
90.1 &
65.4 &
\textbf{70.7} \\
\midrule
Majority Labels$_\text{val}$                &
64.0 &
69.9 &
35.7 &
- &
68.1 &
57.7 &
50.9 &
62.7 &
53.9 &
51.8 &
61.5 &
57.6 \\
\bottomrule
\end{tabular}%
}
\caption{Detailed results for every task in die (Super)GLUE benchmark for the \texttt{strict} and \texttt{strict-small} track.}
\label{tab:results_strict_small}
\end{table*}
}

{\renewcommand{\arraystretch}{1.15}
\begin{table*}[]
\resizebox{\textwidth}{!}{%
\centering
\begin{tabular}{lcccccc}
\toprule
Model & Hypernym & QA congruence (easy) & QA congruence (tricky) & Subj.-Aux. Inversion & Turn Taking & Average \\
\midrule

\multicolumn{7}{@{}c@{}}{\vspace{0.4em}\raisebox{-0.4em}{\footnotesize{Strict-small Track (10M Words)}}} \\
BabyLlama & 49.6 & 54.7 & 41.2 & 86.0 & 66.1 & 59.5 \\
LTG-BERT & 54.2 & 62.5 & 49.1 & 79.9 & 58.2 & \textbf{60.8} \\
\midrule
BabyHGRN & 49.8 & 56.2 & 37.6 & 89.6 & 59.6 & 58.6 \\
\midrule
\multicolumn{7}{@{}c@{}}{\vspace{0.4em}\raisebox{-0.4em}{\footnotesize{Strict Track (100M Words)}}} \\
BabyLlama & 45.6 & 56.2 & 44.8 & 83.9 & 72.5 & 60.6 \\
LTG-BERT & 55.0 & 75.0 & 53.3 & 87.5 & 61.4 & \textbf{66.5} \\
\midrule
BabyHGRN & 48.6 & 64.1 & 35.8 & 84.9 & 59.3 & 58.5 \\
\bottomrule
\end{tabular}
}
\caption{Detailed results for the BLiMP-Supplement benchmark for the \texttt{strict} and \texttt{strict-small} track.}
\label{tab:model_comparison}
\end{table*}
}

{\renewcommand{\arraystretch}{1.15}
\begin{table*}[htbp]
\resizebox{\textwidth}{!}{%
\centering
\begin{tabular}{lcccccccccccc}
Model & \rotatebox{90}{Agent Properties} & \rotatebox{90}{Material Dynamics} & \rotatebox{90}{Material Properties} & \rotatebox{90}{Physical Dynamics} & \rotatebox{90}{Physical Interactions} & \rotatebox{90}{Physical Relations} & \rotatebox{90}{Quantitative Properties} & \rotatebox{90}{Social Interactions} & \rotatebox{90}{Social Properties} & \rotatebox{90}{Social Relations} & \rotatebox{90}{Spatial Relations} & \rotatebox{90}{Macroaverage} \\
\midrule
\multicolumn{13}{@{}c@{}}{\vspace{0.4em}\raisebox{-0.4em}{\footnotesize{Strict-small Track (10M Words)}}} \\
BabyLlama & 50.5 & 51.7 & 49.4 & 54.2 & 50.4 & 50.6 & 53.5 & 50.7 & 50.3 & 49.8 & 46.7 & 50.7 \\
LTG-BERT & 50.2 & 51.0 & 45.3 & 42.5 & 49.1 & 51.0 & 48.1 & 51.7 & 53.4 & 50.6 & 45.3 & 48.9 \\
\midrule
BabyHGRN & 50.1 & 50.9 & 50.6 & 55.0 & 50.7 & 50.4 & 51.3 & 54.1 & 51.2 & 50.3 & 49.8 & \textbf{51.3} \\

\midrule
\multicolumn{13}{@{}c@{}}{\vspace{0.4em}\raisebox{-0.4em}{\footnotesize{Strict Track (100M Words)}}} \\
BabyLlama & 50.1 & 55.5 & 50.0 & 57.5 & 51.4 & 50.5 & 56.7 & 52.7 & 49.7 & 50.0 & 49.0 & \textbf{52.1} \\
LTG-BERT & 50.1 & 55.8 & 50.6 & 58.3 & 48.9 & 50.9 & 53.8 & 51.4 & 50.8 & 53.8 & 49.2 & 51.9 \\
\midrule
BabyHGRN & 50.2 & 52.5 & 51.8 & 49.2 & 51.4 & 50.6 & 54.5 & 51.4 & 57.0 & 49.7 & 49.6 & 51.6 \\

\bottomrule
\end{tabular}%
}
\caption{Detailed results for the EWoK benchmark for the \texttt{strict} and \texttt{strict-small} track.}
\label{tab:ewok_scores}
\end{table*}
}
\end{document}